\documentclass{article}

\usepackage[final]{style}

\usepackage[utf8]{inputenc} 
\usepackage[T1]{fontenc}    
\usepackage{hyperref}       
\usepackage{url}            
\usepackage{booktabs}       
\usepackage{amsfonts}       
\usepackage{nicefrac}       
\usepackage{microtype}      
\usepackage{graphicx}
\usepackage{enumitem}
\usepackage{color}
\usepackage{amsmath}
\usepackage{amssymb}
\usepackage{overpic}
\usepackage{wrapfig}
\usepackage{tabularx,ragged2e,booktabs,caption}
\usepackage{siunitx}

\newcolumntype{L}{>{\centering\arraybackslash}m{1.6cm}}

\title{Listen, Interact and Talk: Learning to Speak via Interaction}

\author{
	Haichao Zhang, Haonan Yu, and Wei Xu\\
	Baidu Research - Institue of Deep Learning\\
	Sunnyvale, CA 94089\\
	\texttt{\{zhanghaichao,haonanyu,xuwei06\}@baidu.com}\\
}

\begin{document}

\maketitle

\begin{abstract}
One of the long-term goals of artificial intelligence is to build an agent that can  communicate intelligently with human in natural language. Most existing work on natural language learning relies heavily on training over a pre-collected dataset with annotated labels, leading to an agent that essentially captures the statistics of the fixed external training data. As the training data is essentially a static snapshot representation of  the knowledge from the annotator, the agent trained this way is limited in adaptiveness and generalization of its behavior. Moreover, this is very different from the language learning process of humans, where language is acquired  during communication by taking speaking action and learning from the consequences of speaking action in an interactive manner. This paper presents an interactive setting for grounded natural language learning, where an agent learns natural language by interacting with a teacher and learning from feedback, thus learning and improving language skills while taking part in the conversation. To achieve this goal, we propose a model which incorporates both imitation and reinforcement by leveraging jointly sentence and reward feedbacks from the teacher. Experiments are conducted to validate the effectiveness of the proposed approach.

\end{abstract}

\section{Introduction}\label{sec:Intro}

Natural language is the one of the most natural form of communication for human, and therefore it is of great value for an intelligent  agent to be able to leverage natural language as the channel to communicate with human as well.
Recent progress on natural language learning mainly relies on supervised training with large scale training data, which typically requires a huge amount of human labor for annotating. While promising performance has been achieved in many specific applications regardless of the labeling effort, this is very different from how humans learn.
Humans act upon the world and learn from the consequences of their actions~\citep{BFSkinner}.
For mechanical actions such as movement, the consequences mainly follow geometrical and mechanical principles, while for language, humans act by speaking and the consequence is typically response in the form of verbal and other behavioral feedbacks (\emph{e.g.}, nodding) from conversation partners. These feedbacks typically contain  informative signal on how to improve the language skills in  subsequent conversions and play an important role in human’s language acquisition process~\citep{rl_lang_learning,nature_lang,DBLL}.

The language acquisition process of a baby is both impressive as a manifestation of human intelligence and inspiring for designing novel  settings and algorithms for computational language learning.
For example,  baby interacts with people and learn through mimicking and feedbacks~\citep{nature_lang,BFSkinner}. 
For learning to speak, baby initially performs verbal action by mimicking his conversational parter (\emph{e.g.} parent) and masters the skill of generating a word (sentence). He could also possibly pick up the association of a word with a visual image when his parents saying ``\emph{this is apple}'' while pointing to an apple or an image of it.
Later, one can ask the baby question like ``\emph{what is this}'' while pointing to an  object, and provides the correct answer if the baby doesn't respond or responds incorrectly, which is typical in the initial stage.
One can also provide at the same time a verbal confirmation (\emph{e.g.} ``\emph{yes/no}'') with a nodding/smile/kiss/hug when he answers correctly as a form of encouragement feedback.
From a baby's perspective, the way to learn the language is by making verbal utterances to parent and adjusting his verbal behavior according to the corrections/confirmation/encouragement from parent.

\begin{figure}[t]
	\hspace*{-0.5cm} 
	\begin{overpic}[viewport = 39 39 242 241, clip, width = 3cm]{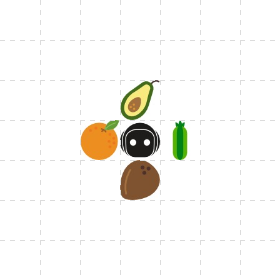}
		\put(90,101){\sffamily \scriptsize {\textcolor{magenta}{{--------question-answer-feedback--------}}}}
		\put(90,92){\sffamily \scriptsize {\textcolor{black}{{Teacher: \emph{what is on the north}}}}}
		\put(90,84){\sffamily \scriptsize {\textcolor{black}{{Learner: \emph{ on .  cabbage yes east}}}}}
		\put(90,76){\sffamily \scriptsize {\textcolor{black}{{Teacher: \emph{on the north is avocado}} {\color{red}[$-$]}}}}
		
		\put(90,65){\sffamily \scriptsize {\textcolor{magenta}{{--------statement-repeat-feedback--------}}}}		
		\put(90,56){\sffamily \scriptsize {\textcolor{black}{{Teacher:  \emph{on the west is orange}}}}}
		\put(90,48){\sffamily \scriptsize {\textcolor{black}{{Learner: \emph{on the west is apple}}}}}
		\put(90,40){\sffamily \scriptsize {\textcolor{black}{{Teacher: \emph{no orange is on the west} {\color{red}[$-$]}}}}}
		
		\put(90,29){\sffamily \scriptsize {\textcolor{magenta}{{-------learner-statement-feedback--------}}}}			
		\put(90,20){\sffamily \scriptsize {\textcolor{black}{{Teacher: .}}}}
		\put(90,12){\sffamily \scriptsize {\textcolor{black}{{Learner: \emph{cucumber is on the east}}}}}
		\put(90,4){\sffamily \scriptsize {\textcolor{black}{{Teacher: \emph{cucumber is on the east} {\color{blue}[$+$]}}}}}
		\put(45,-3){\sffamily \scriptsize {\textcolor{black}{{(a) Training}}}}
	\end{overpic}
	\hspace{4cm}
	\begin{overpic}[viewport = 39 39 242 241, clip, width = 3cm]{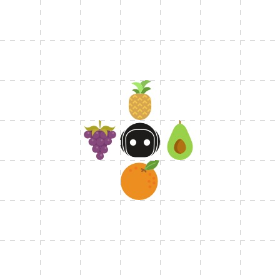}
		\put(90,101){\sffamily \scriptsize {\textcolor{magenta}{{------compositional-generalization------}}}}
		\put(90,92){\sffamily \scriptsize {\textcolor{black}{{Teacher: \emph{what is on the east}}}}}
		\put(90,83){\sffamily \scriptsize {\textcolor{black}{{Learner: \emph{avocado is on the east}}}}}
		\put(90,74){\sffamily \scriptsize {\textcolor{black}{{Teacher: \emph{yes on the east is avocado}}}}}
		
		\put(90,62){\sffamily \scriptsize {\textcolor{black}{{Teacher: \emph{where is avocado}}}}}
		\put(90,53){\sffamily \scriptsize {\textcolor{black}{{Learner: \emph{avocado is on the east}}}}}
		\put(90,44){\sffamily \scriptsize {\textcolor{black}{{Teacher: \emph{yes avocado is on the east}}}}}
		
		\put(90,33){\sffamily \scriptsize {\textcolor{magenta}{{----------knowledge-transferring---------}}}}		
		\put(90,23){\sffamily \scriptsize {\textcolor{black}{{Teacher: \emph{what is on the south}}}}}
		\put(90,14){\sffamily \scriptsize {\textcolor{black}{{Learner: \emph{on the south is orange}}}}}
		\put(90,5){\sffamily \scriptsize {\textcolor{black}{{Teacher: \emph{yes orange is on the south}}}}}			
		\put(45,-3){\sffamily \scriptsize {\textcolor{black}{{(b) Testing}}}}
	\end{overpic}
	\caption{\textbf{Interactive language learning example.} (a) During training, teacher interacts in natural language with learner about objects.
		The interactions are in the form of (1)~question-answer-feedback, (2)~statement-repeat-feedback, and (3)~statement from learner and then feedback from teacher. 
		Certain forms of interactions may be excluded for certain set of object-direction combinations or objects (referred to as \emph{inactive combinations/objects}) during training. For example,  
		the combination of \{\emph{avocado}, \emph{east}\} does not appear in question-answer sessions;
		the object \emph{orange} never appears in question-answer sessions but only in statement-repeat sessions.
		Teacher provides both sentence feedback as well as reward signal (denoted as {\color{blue}[$+$]} and {\color{red}[$-$]} in the figure).
		(b) During testing, teacher can ask question about objects around, including questions involving \emph{inactive combinations/objects} that have never been asked before, \emph{e.g.}, questions about the combination of \{\emph{avocado}, \emph{east}\} and questions about \emph{orange}.
		This  testing setup involves \emph{compositional generalization} and \emph{knowledge transferring} settings and is used for evaluating the proposed approach (\emph{c.f. Section~\ref{sec:Exp}}). 
	} 
	\label{fig:example}
	\vspace{-0.2in}	
\end{figure}
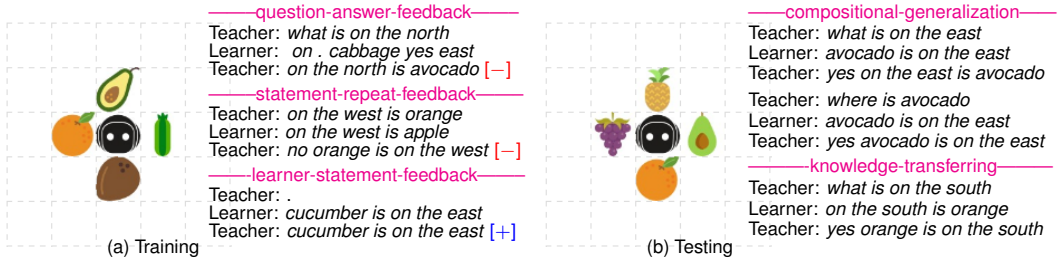

This example illustrates that the language learning process is inherently  \emph{interactive}, a property which is potentially difficult to be captured by a static dataset as used in the conventional supervised learning setting.  
Inspired by baby's language learning process, we present a novel interactive setting for grounded natural language learning,  where the teacher and the learner can interact with each other in natural languages as shown in Figure~\ref{fig:example}. 
In this setting, there is no direct  supervisions to guide the behavior of the learner as in the  supervised learning setting. Instead, the 
learner has to \emph{act in order to learn}, \emph{i.e.},  engaging in the conversation with currently acquired speaking skills to obtain feedbacks from the dialogue partner, which provide learning signals   for further improvement on the conversation skills.

To leverage the feedbacks for learning, it is tempting to mimic the teacher directly (\emph{e.g.},  using a language model). While this is a viable approach for learning how to speak, the agent trained by pure imitation is not necessarily able to converse adaptively within context due to the negligence of the reinforcement signal.
An example is that it is hard to make a successful conversation with a well-trained parrot, which is only good at mimicking.
The reason is that the learner is mimicking from a third person perspective~\citep{third_person},  mimicking the teacher who is conversing with it, thus certain words in the sentences from the teacher such as ``\emph{yes/no}'' and ``\emph{you/I}'' might need to be removed/adapted due to the change of perspective from teacher to learner. This cannot be achieved with  imitation only. 
On the other hand, it is also challenging to generate appropriate conversational actions using purely the reinforcement signal without imitation.
The fundamental reason is the inability of speaking, thus the probability of generating a sensible sentence by randomly uttering is low, let alone that of a proper one.
This is exemplified by the fact that babies don't fully develop their language capabilities without the ability to hear, which is one of the most important channels for language-related imitation.

In this paper, we propose a \emph{joint imitation and reinforcement} approach for interactive language learning. The proposed approach leverages both verbal and encouragement feedbacks from the teacher for joint learning, thus overcoming the difficulties encountered with either only  imitation or reinforcement.
The contributions of this paper can be therefore summarized as the following:
\vspace{-2.5mm}
\begin{itemize}[leftmargin=10pt]
	\itemsep-0.08em
	\item We present a novel human-like interaction-based grounded language learning setting where language is learned by interacting with the environment (teacher) in natural language. 
	\item We present a grounded natural language learning approach under the interactive setting  by leveraging feedbacks from the teacher during interaction through joint imitation and reinforcement.
\end{itemize}
\vspace{-2.5mm}
To the best of our knowledge, this is the first work on using imitation and reinforcement jointly for grounded natural language learning in an interactive setting.

The remainder of the paper is structured as follows.  In Section~\ref{sec:related_works}, we make a brief review of  related work on natural language  learning.
Section~\ref{sec:model} introduces the formulation of the interaction-based natural language learning problem, followed with detailed explanation of the proposed approach.
Experiments are carried out in Section~\ref{sec:Exp} to show the language learning ability of the proposed approach in the interactive setting. Finally, we conclude the paper in Section~\ref{sec:con}.

\vspace{-0.1in}
\section{Related Work}
\vspace{-0.05in}
\label{sec:related_works}
%
\vspace{-0.1in}
Deep network based language learning has received great success recently and has been applied in different applications, for example,  machine translation~\citep{Seq2Seq}, image captioning/visual question answering~\citep{mao2014deep, show_and_tell, VQA} and dialogue response generation~\citep{Neural_conversation, lang_generation}.
For training, a large amount of training data containing source-target pairs is needed, typically requiring a significant amount of efforts to collect.
This setting essentially captures the statistics of the training data and does not respect the interactive nature of language learning thus is very different from how humans learn.

While conventional language model is trained in a supervised way, 
there are some recent works using  reinforcement learning for training.  These works mainly target at the problem of tuning the performance of a language model pre-trained in a supervised way according to a specific reward function which is either directly the evaluation metric such as standard BLEU core~\citep{RL_Seq_ICLR15, AC_Seq}, manually designed function~\citep{RL_Dialogue} or metric learned in an adversarial setting~\citep{SeqGAN, Adversarial_Dialogue}, which is non-differentiable, leading to the usage of reinforcement learning.
Different from them, our main focus is on the possibility of language learning in an interactive setting and required model designs, rather than optimizing a particular model output towards a specific evaluation metric.

There are some recent works on learning to communicate~\citep{RL_Com, BP_Com} and the emergence of language~\citep{Multi_Agent_Lan,Emergence_Lan}. The emerged language need to be interpreted via post-processing~\citep{Emergence_Lan}.
Differently, we aim to achieve natural language learning from both  perspectives of understanding and generation (\emph{i.e.}, \emph{speaking}), thus the speaking action of the agent is readily understandable without any post-processing.
There are also works on dialogue learning using a guesser/responser setting where the guesser tries to achieve the final goal (\emph{e.g.}, classification/localization) by collecting additional information through asking questions to the responser~\citep{Dialogue_Deepmind, visdial_rl}. 
These works try to optimize the question to be asked in order to help the guesser to achieve the final guessing goal.
Thus the focus is very different from our goal of language learning through interactions with teacher.

Our work is also related to reinforcement learning based control with natural language action space~\citep{Lan_Action_Space} in the sense that our model also outputs action in natural language space.
We also shares similar motivation with~\citep{DBLL, D_interaction}, 
where language learning through textual dialogue has been explored.
However, in these works~\citep{Lan_Action_Space, DBLL, D_interaction}
a set of candidate sequences is provided and the action required is selecting one from the candidate set, thus is essentially a \emph{discrete control} problem.
In contrast, our model achieves sentence generation through control in a \emph{ continuous space},  with a potentially infinite sized action space consisting of all possible sequences.

\vspace{-0.15in}
\section{Interaction-based Language Learning}  \label{sec:model}
\vspace{-0.15in}
We will introduce the proposed  interaction-based natural language learning approach in this section.
The goal is to  design a learning agent\footnote{We use the term \emph{agent} interchangeably with \emph{learner} according to context in the paper.} that can learn to converse by interacting with the teacher, which can be either a virtual teacher or a human (\emph{c.f.} Figure~\ref{fig:example}$\sim$\ref{fig:NN}).
At time step $t$, according to a visual image $\mathbf{v}$, teacher generates a sentence $\mathbf{w}^t$ which can be a question (\emph{e.g.}, ``\emph{what is on the east}'', ``\emph{where is apple}''), a statement (\emph{e.g.}, ``\emph{banana is on the north}''), or an empty sentence (denoted as ``.''). The learner takes teacher's sentence $\mathbf{w}^t$  and the visual content  $\mathbf{v}$, and produces a sentence response  $\mathbf{a}^t$ to the teacher. The teacher will then provide feedbacks to the learner according to its response in the form of both sentence  $\mathbf{w}^{t+1}$ and reward  $r^{t+1}$.
The sentence $\mathbf{w}^{t+1}$ represents verbal feedback from teacher (\emph{e.g.}, ``\emph{yes on the east is cherry}'', ``\emph{no apple is on the east}'') and $r^{t+1}$ models the non-verbal confirmative feedback such as nodding/smile/kiss/hug, which also appears naturally during interaction.
The problem is therefore to design a model that can learn grounded natural language from teacher's sentences and reward feedbacks.
While it might looks promising to formulate the problem as supervised training by learning from the subset of sentences from teacher with only positive rewards, this approach won't work because of the difficulties due to the changed of perspective~\citep{third_person} as mentioned earlier.  
Our formulation of the problem as well as the details of the proposed approach  are presented in the sequel.

\vspace{-0.1in}
\subsection{Problem Formulation}
\begin{figure}[t]
	\hspace{0.15in}
	\begin{overpic}[viewport = 1 -10 780 450, clip, width = 10cm]{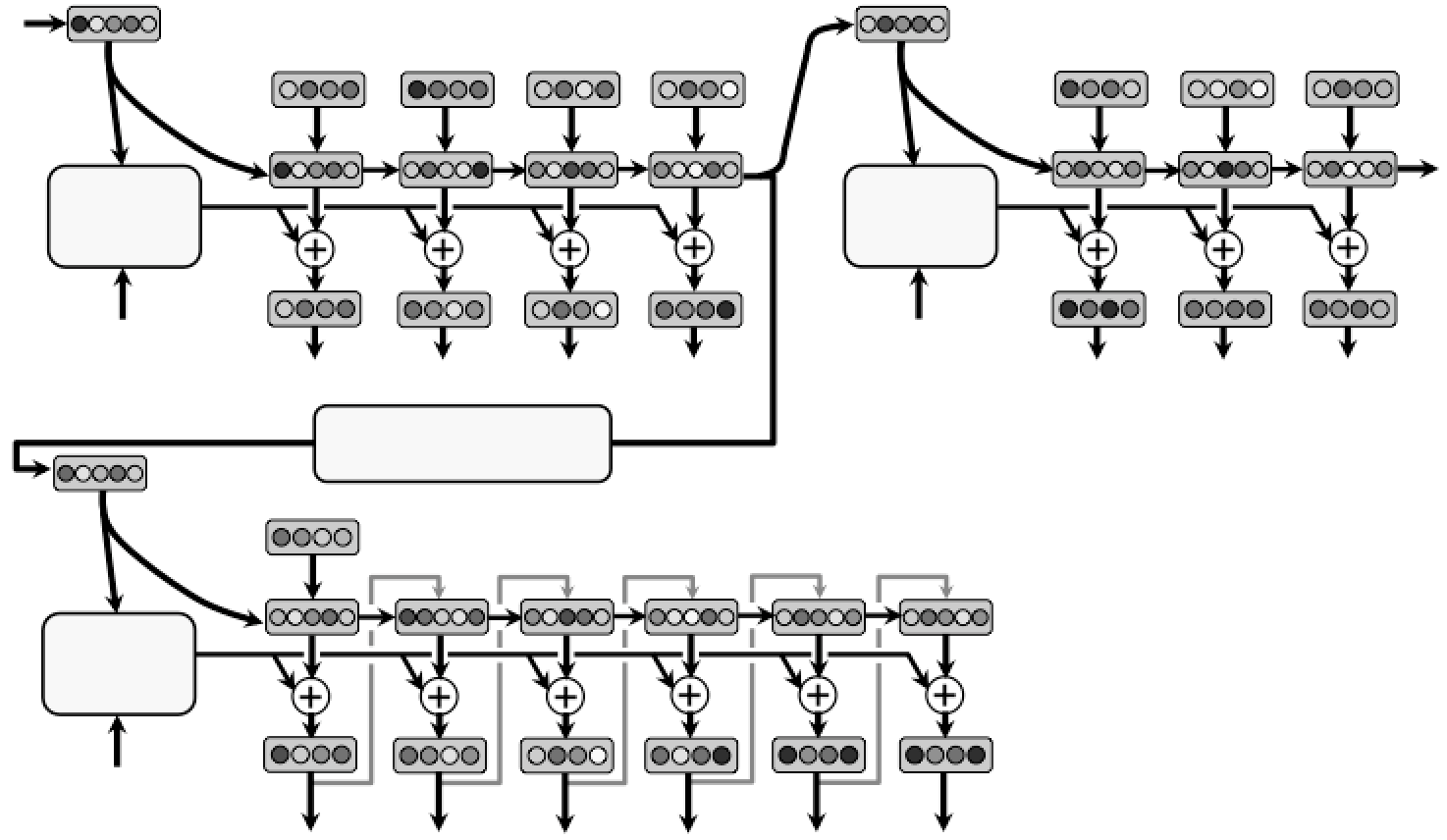}
		\put(4.2,1.3){\fbox{\includegraphics[viewport = 75 75 205 205, clip, width = 0.55cm]{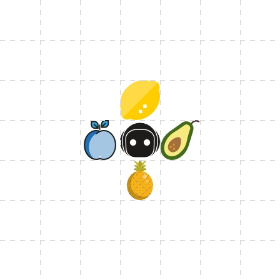}}}
		\put(4.5,30){\fbox{\includegraphics[viewport = 75 75 205 205, clip, width = 0.55cm]{Visualization_color_img_3}}}
		\put(56.6,29.8){\fbox{\includegraphics[viewport = 75 75 205 205, clip, width = 0.55cm]{Visualization_color_img_3}}}
		\put(31.5,55){{{\textcolor{black}{\scalebox{.8}{$\mathbf{w}^{t}$}}}}}
		\put(77,55){{{\textcolor{black}{\scalebox{.8}{$\mathbf{w}^{t+1}$}}}}}
		\put(31,-2){{{\textcolor{black}{\scalebox{.8}{$\mathbf{a}^{t}$}}}}}				
		\put(2,32){{{\textcolor{black}{\scalebox{.8}{$\mathbf{v}$}}}}}
		\put(54.5,32){{{\textcolor{black}{\scalebox{.8}{$\mathbf{v}$}}}}}
		\put(2,2){{{\textcolor{black}{\scalebox{.8}{$\mathbf{v}$}}}}}
		
		\put(-5,55){{{\textcolor{black}{\scalebox{.8}{$\mathbf{h}_{\rm last}^{t-1}$}}}}}
		\put(47,55){{{\textcolor{black}{\scalebox{.8}{$\mathbf{h}_{\rm last}^{t}$}}}}}
		\put(7.5,57){{{\textcolor{black}{\scalebox{0.8}{$t \longmapsto$}}}}}
		\put(59.5,57){{{\textcolor{black}{\scalebox{0.8}{$t+1 \longmapsto$}}}}}

		\put(9,51){{{\textcolor{black}{\scalebox{.8}{$\mathbf{h}_{0}^{t}$}}}}}	
		\put(60,51){{{\textcolor{black}{\scalebox{.8}{$\mathbf{h}_{0}^{t+1}$}}}}}				
		\put(-4,34){\sffamily \small {\textcolor{black}{{\scalebox{.8}{\rotatebox{90}{Encoding-RNN}}}}}}
		\put(-4,2){\sffamily \small {\textcolor{black}{{\scalebox{.8}{\rotatebox{90}{Action-RNN}}}}}}
		\put(-3,25){\vector(0,1){9}}
		\put(-3,26.5){\vector(0,-1){11}}
		\put(-5,17.6){\sffamily \scriptsize {\textcolor{black}{{\scalebox{.8}{\rotatebox{90}{share parameter}}}}}}
				
		\put(17,52.5){{\sffamily \scalebox{.6} {\textcolor{black}{\emph{<bos>}}}}}
		\put(26,52.5){{\sffamily \scalebox{.6} {\textcolor{black}{\emph{where}}}}}
		\put(36,52.5){{\sffamily \scalebox{.6} {\textcolor{black}{\emph{is}}}}}
		\put(42,52.5){{\sffamily \scalebox{.6} {\textcolor{black}{\emph{apple}}}}}
					
		\put(17,31){{\sffamily \scalebox{.6} {\textcolor{black}{\emph{where}}}}}
		\put(27.5,31){{\sffamily \scalebox{.6} {\textcolor{black}{\emph{is}}}}}
		\put(34,31){{\sffamily \scalebox{.6} {\textcolor{black}{\emph{apple}}}}}
		\put(42,31){{\sffamily \scalebox{.6} {\textcolor{black}{\emph{<eos>}}}}}

		\put(69,52.5){{\sffamily \scalebox{.6} {\textcolor{black}{\emph{<bos>}}}}}
		\put(79,52.5){{\sffamily \scalebox{.6} {\textcolor{black}{\emph{no}}}}}
		\put(86,52.5){{\sffamily \scalebox{.6} {\textcolor{black}{\emph{west}}}}}
		\put(70,31){{\sffamily \scalebox{.6} {\textcolor{black}{\emph{no}}}}}
		\put(77.5,31){{\sffamily \scalebox{.6} {\textcolor{black}{\emph{west}}}}}
		\put(85,31){{\sffamily \scalebox{.6} {\textcolor{black}{\emph{<eos>}}}}}		
			
		\put(17.5,22.8){{\sffamily \scalebox{.6} {\textcolor{black}{\emph{<bos>}}}}}
		\put(18,0.2){{\sffamily \scalebox{.6} {\textcolor{black}{\emph{apple}}}}}
		\put(27.5,0.2){{\sffamily \scalebox{.6} {\textcolor{black}{\emph{is}}}}}
		\put(35.5, 0.2){{\sffamily \scalebox{.6} {\textcolor{black}{\emph{on}}}}}
		\put(43,0.2){{\sffamily \scalebox{.6} {\textcolor{black}{\emph{the}}}}}
		\put(50.5,0.2){{\sffamily \scalebox{.6} {\textcolor{black}{\emph{south}}}}}
		\put(58.5,0.2){{\sffamily \scalebox{.6} {\textcolor{black}{\emph{<eos>}}}}}

		\put(24.,26.5){{\sffamily \scalebox{.6} {\textcolor{black}{{controller $f$}}}}}
		\put(13 ,24.5){{\sffamily \scalebox{.8} {\textcolor{black}{{$\mathbf{k}^t$}}}}}
		
		\put(4.6,42.6){{\sffamily \scalebox{.6} {\textcolor{black}{{visual}}}}}
		\put(4,40.4){{\sffamily \scalebox{.6} {\textcolor{black}{{encoder}}}}}	
		
		\put(56.6,42.6){{\sffamily \scalebox{.6} {\textcolor{black}{{visual}}}}}
		\put(56,40.4){{\sffamily \scalebox{.6} {\textcolor{black}{{encoder}}}}}		
		
		\put(4.3,13){{\sffamily \scalebox{.6} {\textcolor{black}{{visual}}}}}
		\put(3.7,11){{\sffamily \scalebox{.6} {\textcolor{black}{{encoder}}}}}	
					
		\put(-2,-2){\sffamily \scriptsize {\textcolor{black}{{(a)}}}}		
		
		\put(68, 3){\includegraphics[viewport = 19 9 360 250, clip, width = 2.8cm]{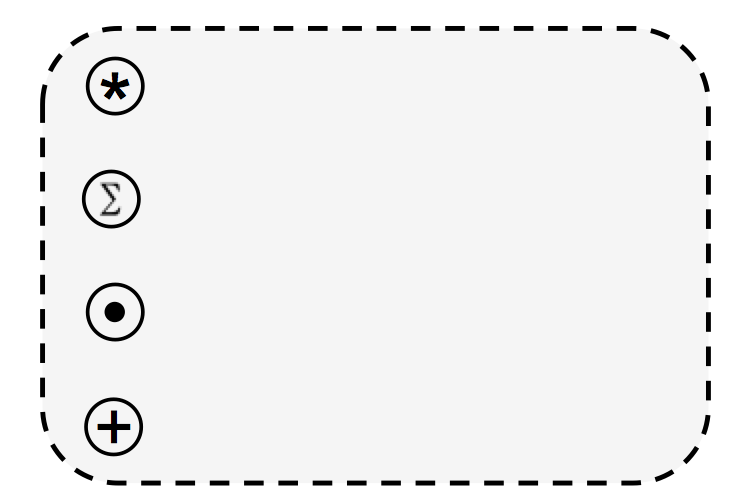}}		
		\put(75,18.5){{\sffamily \scalebox{.6} {\textcolor{black}{{spatial convolution}}}}}
		\put(75,13.8){{\sffamily \scalebox{.6} {\textcolor{black}{{spatial summation}}}}}
		\put(75,9.){{\sffamily \scalebox{.6} {\textcolor{black}{{Hadamard product}}}}}
		\put(75,4.5){{\sffamily \scalebox{.6} {\textcolor{black}{{mix aggregation}}}}}

		\put(102, 33){\includegraphics[viewport = 8.5 10 470 550, clip, width = 3cm]{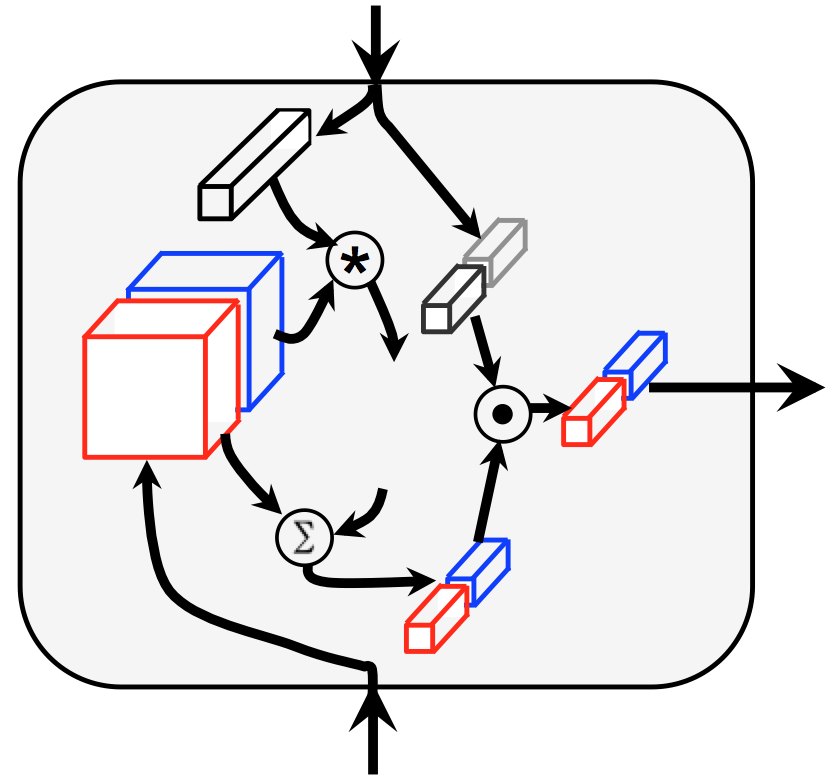}}	
		\put(103.5,52){{{\textcolor{black}{\scalebox{.6}{$V_{\rm att}$}}}}}
		\put(112.,42){{\includegraphics[viewport = 75 75 205 205, clip, width = 0.4cm]{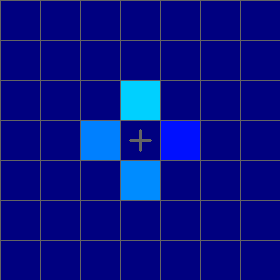}}}
		\put(103.8,37){{\sffamily \scalebox{.4} {\textcolor{black}{{CNN}}}}}
		\put(112,52.5){{\sffamily \scalebox{.4} {\textcolor{black}{{att.}}}}}
		\put(112,51){{\sffamily \scalebox{.4} {\textcolor{black}{{filter}}}}}
		\put(118,51){{\sffamily \scalebox{.5} {\textcolor{black}{{mask}}}}}
		\put(118,38){{\sffamily \scalebox{.4} {\textcolor{black}{{aggregated}}}}}
		\put(118,36.5){{\sffamily \scalebox{.4} {\textcolor{black}{{feature}}}}}
		\put(114.8,33){{{\textcolor{black}{\scalebox{.8}{$\mathbf{v}$}}}}}	
		\put(114.8,56.2){{{\textcolor{black}{\scalebox{.8}{$\mathbf{h}^t_{0}$}}}}}
		\put(104,30.5){\sffamily \scriptsize {\textcolor{black}{{(b) visual encoder}}}}

		\put(102, 2){\includegraphics[viewport = 8.5 5 470 550, clip, width = 3cm]{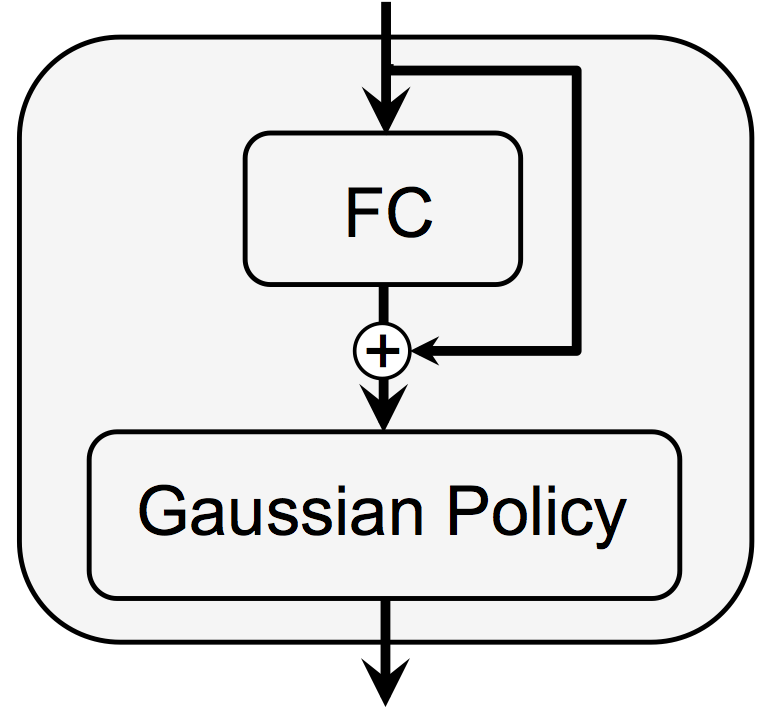}}
		\put(103.5,20){{{\textcolor{black}{\scalebox{.8}{$f$}}}}}
		\put(113.1,25){{{\textcolor{black}{\scalebox{.8}{$\mathbf{h}^t_{\rm last}$}}}}}	
		\put(115,1.5){{{\textcolor{black}{\scalebox{.8}{$\mathbf{k}_t$}}}}}
		\put(121, 13){\sffamily \scriptsize {\textcolor{black}{{\scalebox{.7}{\rotatebox{90}{skip connect.}}}}}}
			
		\put(107,-1.5){\sffamily \scriptsize {\textcolor{black}{{(c) controller}}}}	
	\end{overpic}
	\vspace{0.1in}
	\caption{\textbf{Network structure.} (a) Illustration of the network structure with sample inputs. 
		(b) Visual encoder network $V_{\rm att}(\cdot)$.
	Visual image is encoded by a CNN and spatially aggregated to a vector with an attention map. The attention map is obtained by convolving the feature map from CNN with a spatial filter generated from the initial state $\mathbf{h}^t_0$. A binary mask generated from $\mathbf{h}^t_0$ is applied to the spatially aggregated vector to produce the final visual feature vector.
	At time step $t$, the encoding-RNN takes teacher's sentence (``\emph{where is apple}'') and the visual feature vector from the visual encoder $V_{\rm att}(\cdot)$ as inputs. The last state of the encoding-RNN $\mathbf{h}_{\rm last}^t$ is passed through a controller $f(\cdot)$ to the action-RNN for response generation. Parameters are shared between encoding-RNN and action-RNN.
	During training, the RNN is trained by predicting next words and next sentences.
	(c)~Controller network with a residue control module followed by a Gaussian Policy module (c.f. Sec.~\ref{RL_hRNN}).}
	\label{fig:NN}
	\vspace{-0.2in}
\end{figure}

\vspace{-0.05in}
A response from the agent can be modeled as a sample from a probability distribution over the possible output sequences. Specifically, for one episode, given the visual input $\mathbf{v}$ and  textual input $\mathbf{w}^{1:t}$  from teacher upto time step $t$, the response $\mathbf{a}^t$ from the agent can be generated by sampling from a policy distribution $p_{\theta}^{\rm R}(\cdot)$ of the speaking action:
\begin{eqnarray}
\label{eq:action_generation}
\begin{split}
\mathbf{a}^t &\sim p_{\theta}^{\rm R}(\mathbf{a}|\mathbf{w}^{1:t}, \mathbf{v}).
\end{split}
\end{eqnarray}
The agent \emph{interacts} with teacher by outputting the utterance $\mathbf{a}^t$ and receives the \emph{feedbacks} from teacher at time step $t+1$ as  $\mathcal{F}=\{\mathbf{w}^{t+1}, r^{t+1}\}$. 
$\mathbf{w}^{t+1}$ is in the form of a sentence which represents a verbal confirmation/correction in accordance with $\mathbf{w}^t$ and $\mathbf{a}^t$, with prefixes (\emph{yes/no}) added with a  probability of half (\emph{c.f.} Figure~\ref{fig:example}$\sim$\ref{fig:NN}).
Reward $r^{t+1}$ is a scalar-valued feedback with positive value as encouragement while negative value as discouragement according to the correctness of the agent utterance $\mathbf{a}^t$. 
The task of interaction-based language learning can be stated as 
\emph{learning by conversing with teacher and improving from teacher's feedbacks  $\mathcal{F}$}.
Mathematically, we formulate the problem as the minimization of a cost function as follows:
\begin{eqnarray}
\label{cost_function}
\begin{split}
\mathcal{L}_{\theta} = \mathcal{L}^{\rm I}_{\theta} + \mathcal{L}^{\rm R}_{\theta} =  \underbrace{\mathbb{E}_{S}\Big[ -\textstyle \sum_t\log p_{\theta}^{\rm I}(\mathbf{w}^{t+1}|\mathbf{w}^{1:t}, \mathbf{v})\Big]}_{\rm Imitation} + \underbrace{\mathbb{E}_{p^{\rm R}_{\theta}}  \Big[-\textstyle \sum_t [\gamma]^{t} \cdot r^{t+1}\Big]}_{\rm Reinforce},
\end{split}
\end{eqnarray}
where $\mathbb{E}_{S}(\cdot)$ is the expectation over all the sentence sequences $S$ generated from teacher,
$r^{t+1}$  is the  immediate reward  received at time step $t+1$ after taking speaking action following policy $p^{\rm R}_{\theta}(\cdot)$ at time step $t$ and $\gamma$ is the reward discount factor.  
$[\gamma]^{t}$ is used to denote the exponentiation over $\gamma$ to differentiate it with superscript indexing.
As for both components, the training signal is obtained via \emph{interaction} with the teacher, we termed this task as \emph{interaction}-based language learning. For the imitation part, it  essentially learns from teacher's verbal response $\mathbf{w}^{t+1}$, which can only be obtained as a consequence of its speaking action. For the reinforce part, it learns from teacher's reward signal $r^{t+1}$, which is also obtained after taking the speaking action and received at the next time step. 
The proposed interactive language learning formulation integrates two components which can fully leverage the feedbacks appearing naturally  during conversational interaction:
\vspace{-0.15cm}
\begin{itemize}[leftmargin=10pt]
\item \textbf{Imitation} plays the role of learning a grounded language model by observing teacher's behaviors during conversion with the learner itself. This  enables the learner to have the basic ability to speak within context. The training data here are only the sentences from teacher, without any explicit labeling of ground-truth and is a mixture of expected correct response and others. The way of training is by \emph{predicting the future}. More specifically, the model is predicting the next future word at word level and predicting the next sentence at sentence level. 
Another important point is that it is in effect \emph{third person imitation}~\citep{third_person}, as the learner is imitating the teacher who is conversing with it, rather than another expert student who is conversing with teacher.
\item \textbf{Reinforce}\footnote{Reinforce denotes the module that learns from the reinforcement/encouragement signal throughout the paper and should be differentiated with the REINFORCE algorithm in the literature~\citep{Sutton}.} leverages the confirmative feedbacks from the teacher for learning to converse properly by adjusting the action policy distribution. It enables the learner to use the acquired speaking ability and adapt it according to feedbacks. Here we have the learning signal in the form of reward.  This is analogous to baby's language learning process, who uses the acquired language skills by trial and error with parents and improves according to the encouragement feedbacks. 
\end{itemize}
\vspace{-0.05cm}
Note that while imitation and reinforce are represented as two separate components in Eqn.(\ref{cost_function}), they are tied via parameter sharing in order to fully leverage both forms of training signals.
This form of joint learning is crucial for achieving successful language learning, compared with approaches with only imitation or reinforce which are less effective, as verified by experiments in Section~\ref{sec:Exp}.

\vspace{-0.1in}
\subsection{Approach}
\vspace{-0.1in}
\label{sec:approachl}
A hierarchical Recurrent Neural Network is used for capturing the sequential structure both across sentences and within a sentence~\citep{YuWHYX16, HRNN_dialogue16}, as shown in Figure~\ref{fig:NN}(a). 
At time-step $t$, an encoding-RNN encodes the input sentence $\mathbf{w}^t$ from teacher as well as history information into a state vector $\mathbf{h}_{\rm last}^t$, which is passed through an action controller $f(\cdot)$ to produce a control vector $\mathbf{k}^t$ as  input to the action-RNN for generating the response $\mathbf{a}^t$ to the teacher's sentence. Teacher will generate feedback $\mathcal{F}=\{\mathbf{w}^{t+1}, r^{t+1}\}$ according to both $\mathbf{w}^t$ and $\mathbf{a}^t$. 
In addition to being used as  input to action controller, the state vector is also passed to the next time step and used as the initial state of the encoding-RNN in the next step (\emph{i.e.}, $\mathbf{h}_{0}^{t+1} \triangleq \mathbf{h}_{\rm last}^t$) for learning from $\mathbf{w}^{t+1}$, thus forming another level of recurrence at the scale of time steps.

\vspace{-0.05in}
\subsubsection{Imitation with Hierarchical-RNN-based Language Modeling}
\vspace{-0.05in}
The teacher's way of speaking provides a source for the learner to mimic. One way to learn from this source of information is by predictive imitation. 
Specifically, for a particular episode, we can represent the  probability of the next sentence $\mathbf{w}^{t+1}$ conditioned on the previous sentences $\mathbf{w}^{1:t}$ and current image $\mathbf{v}$ as
\begin{eqnarray}
\label{I_prob}
\begin{split}
p_{\theta}^{\rm I}(\mathbf{w}^{t+1}| \mathbf{w}^{1:t}, \mathbf{v})
=p_{\theta}^{\rm I}(\mathbf{w}^{t+1}| \mathbf{h}^{t}_{\rm last}, \mathbf{v})
={\textstyle \prod_{i}}  \; p_{\theta}^{\rm I}(w^{t+1}_i|w_{1:i-1}^{t+1}, \mathbf{h}^{t}_{\rm last}, \mathbf{v}),
\end{split}
\end{eqnarray} 
where $\mathbf{h}^{t}_{\rm last}$ is the last state of RNN at time step $t$ as the summarization of $\mathbf{w}^{1:t}$ (\emph{c.f.} Figure~\ref{fig:NN}), and
$i$ indexes words within a sentence.
It is natural to model the probability of the $i$-th word in the $t\!\!+\!\!1$-th sentence with an RNN  as well, where the sentences up to $t$ and words up to $i$ within the $t\!+\!1$-th sentence we conditioned upon is captured by a fixed-length hidden state vector as $\mathbf{h}_i^{t+1} = {\rm RNN} (\mathbf{h}_{i-1}^{t+1}, {w}_i^{t+1})$, thus
\begin{eqnarray}
\begin{split}
p_{\theta}^{\rm I}(w^{t+1}_i|w_{1:i-1}^{t+1}, \mathbf{h}^{t}_{\rm last}, \mathbf{v}) &= {\rm softmax} (\mathbf{W}_{\rm h}\mathbf{h}_i^{t+1}  + \mathbf{W}_{\rm v} V_{\rm att}(\mathbf{v}, \mathbf{h}_{0}^{t+1}) + \mathbf{b}),
\end{split}
\end{eqnarray}  
where $\mathbf{W}_{\rm h}$, $\mathbf{W}_{\rm v}$ and $\mathbf{b}$ denote the transformation weight and bias parameters respectively.
$V_{\rm att}(\cdot)$ denotes the visual encoding network with 
spatial attention incorporated as shown in Figure~\ref{fig:NN}(b).
$V_{\rm att}(\cdot)$ takes the initial RNN state $\mathbf{h}^t_0$ and visual image $\mathbf{v}$ as input.
The visual image is first encoded by a CNN to obtain the visual feature map (red cube in Figure~\ref{fig:NN}(b)). The visual feature map is appended with another set of maps with learnable parameters for encoding the directional information (blue cube in Figure~\ref{fig:NN}(b)). This set of feature maps is spatially aggregated to a vector with an attention map, which is obtained by convolving the feature map with a spatial filter generated from the initial RNN state. An attention mask for emphasizing visual or directional features is generated from $\mathbf{h}^t_0$  and is applied to the spatially aggregated vector to produce the final feature vector.
The initial state of the  encoding-RNN is the last state of the previous RNN, \emph{i.e.}, $\mathbf{h}_0^{t+1} = \mathbf{h}_{\rm last}^{t}$ and $\mathbf{h}_0^0 = \mathbf{0}$.

The language model trained this way will have the basic ability of producing a sentence conditioned on the input.
Therefore, when connecting an encoding-RNN with action-RNN directly, \emph{i.e.}, inputing the last state vector from encoding-RNN into action-RNN as the initial state, the learner will have the ability to generate a sentence by mimicking the way teacher speaks, due to parameter sharing. However, this basic ability of speaking is not enough for the learner to converse properly with teacher, which requires the incorporation of reinforcement signals as detailed in the following section. 

\vspace{-0.1in}
\subsubsection{Learning via Reinforcement for Sequence Actions}
\vspace{-0.10in}
\label{RL_hRNN}
An agent generates an action according to $p_{\theta}^{\rm R}(\mathbf{a}|\mathbf{w}^{1:t}, \mathbf{v})$.
As sentences $\mathbf{w}^{1:t}$ can be summarized as the last RNN state $\mathbf{h}^t_{\rm last}$, the action policy distribution can be represented as $p_{\theta}^{\rm R}(\mathbf{a}|\mathbf{h}^t_{\rm last}, \mathbf{v})$.
To leverage the language skill that is simultaneously learned from imitation, we can generate the sentence using a language model shared with imitation, but with a modulated conditional signal via a controller network $f(\cdot)$ as follows (\emph{c.f.} Figure~\ref{fig:NN}(a, c))
\begin{eqnarray}
\label{eq:action_prob}
\begin{split}
p_{\theta}^{\rm R}(\mathbf{a}^{t}|\mathbf{h}^t_{\rm last}, \mathbf{v}) = p_{\theta}^{\rm I}(\mathbf{a}^{t}\triangleq\mathbf{w}^{t+1}|f(\mathbf{h}^t_{\rm last}), \mathbf{v}).
\end{split}
\end{eqnarray}
The reason for incorporating a controller $f(\cdot)$ for modulation is that 
the basic language model only offers the learner the ability to generate a sentence, but not necessarily the ability to respond correctly, or to answer a question from teacher properly.
Without any additional module, the agent's behaviors would be the same as those from teacher because of parameter sharing, thus agent cannot learn to  speak correctly in an adaptive manner by leveraging the feedbacks from teacher.

Controller $f(\cdot)$ is a composite function with two components:
(1) a residue structured network for transforming the encoding vector $\mathbf{h}^t_{\rm last}$ in order to modify the behavior; (2) a Gaussian policy module for generating a control vector from a Gaussian distribution conditioned on the transformed encoding vector from the residue control network as a form of exploration. In practice, we also incorporate a gradient-stopping layer between the controller and its input, to encapsulate all the modulation ability within the controller. 
\vspace{-0.1in}
{\flushleft\textbf{Residue Control.}} 
The action controller should have the property that it can pass the input vector to the next module unmodified when desirable while can modify the content of the input vector otherwise. Therefore, a residue structured network is one design satisfying this requirement, with a content modifying vector added to the original input vector (\emph{i.e.}, skip connection) as follows 
\begin{eqnarray}
\label{fc_res}
\begin{split}
\mathbf{c} = {\tau}(\mathbf{h}) + \mathbf{h},
\end{split}
\end{eqnarray}
where $\tau(\cdot)$ is a content transformation net and $\mathbf{c}$ is the generated control vector. 
The reason for including a skip connection is that it offers the ability to leverage the language model simultaneously learned via imitation for generating sensible sentences and the transformation net ${\tau}(\cdot)$ includes learnable parameters for adjusting the behaviors via interactions with the environment and feedbacks from teacher. We implement $\tau(\cdot)$ as two  fully-connected layers with ReLU activation.
\vspace{-0.1in}
{\flushleft \textbf{Gaussian Policy.}}  
Gaussian policy net models the output vector as a Gaussian distribution conditioned on the input vector. 
It takes the generated control vector  $\mathbf{c}$ as input and produces a vector $\mathbf{k}$ which is used as the initial state of the action-RNN. 
The Gaussian policy is modeled as follows:
\begin{eqnarray}
\label{gaussian_policy}
\begin{split}
p^{\rm R}_{\theta}(\mathbf{k} | \mathbf{c}) 
= \mathcal{N}(\mathbf{c}, \mathbf{\Gamma}^T\mathbf{\Gamma}), \; \mathbf{\Gamma} = \mathrm{diag}[\gamma(\mathbf{c})].
\end{split}
\end{eqnarray}
The incorporation of Gaussian policy introduces stochastic unit into the network, thus back-propagation cannot be applied directly. We therefore use policy gradient algorithm for optimization~\citep{Sutton}.
where $\gamma(\cdot)$ is a sub-network for estimating the standard derivation vector and is implemented using a fully-connected layer with ReLU activation.\footnote{In practice, we add a small value ($0.01$) to $\gamma(\mathbf{c})$ as a constrain of the minimum standard deviation.}
The vector $\mathbf{k}$ generated from the controller is then used as the initial state of action-RNN and the sentence output is generated using beam search (\emph{c.f.} Figure~\ref{fig:NN}(a)).
For the reward $r^{t+1}$ in Eqn.(\ref{cost_function}), we introduce a baseline for reducing variance as $r^{t+1} - V_{\upsilon}(\mathbf{v})$, where $V_{\upsilon}(\cdot)$ represents the value network with parameter vector $\upsilon$ and is estimated by adding to $\mathcal{L}^{\rm R}$ an additional value network cost $\mathcal{L}^{\rm V}$ as follows
\begin{eqnarray}
\begin{split}
\mathcal{L}^{\rm V} = \mathbb{E}_{p^{\rm R}_{\theta}} \big(r^{t+1} + \lambda V_{\upsilon^{-}}(\mathbf{v}^{t+1}) - V_{\upsilon}(\mathbf{v}^{t})\big)^2,
\end{split}
\end{eqnarray}
where  $\upsilon$ denotes the set of parameters in the value network and $V_{\upsilon^{-}}(\cdot)$ denotes the target version of the value network, whose parameter vector $\upsilon^{-}$ is periodically copied from the training version~\citep{DQN}. 

\vspace{-0.15in}
\subsection{Training}
\vspace{-0.1in}
Training involves optimizing the stochastic policy by using the teacher's feedback $\mathcal{F}$ as a training signal, obtaining a set of optimized parameters by considering jointly imitation and reinforcement as shown in Eqn.(\ref{cost_function}).
Stochastic gradient descend is used for training the network.
For $\mathcal{L}^{\rm I}$ from imitation module, we have its gradient as:
\begin{eqnarray}
\label{UL_gradient}
\begin{split}
\nabla_{\theta} \mathcal{L}^{\rm I}_{\theta} = - \mathbb{E}_{S} [\nabla_{\theta} \textstyle \sum_t \log p_{\theta}^{\rm I}(\mathbf{w}^{t+1}|\mathbf{w}^{1:t}, \mathbf{v})].
\end{split}
\end{eqnarray}
Using policy gradient theorem~\citep{Sutton}, we have the following gradient for the reinforce module: 
\begin{eqnarray}
\label{gaussian_policy_gradient}
\begin{split}
\nabla_{\theta} \mathcal{L}^{\rm R}_{\theta}  = -\mathbb{E}_{p^{\rm R}_{\theta}}\big[ [\nabla_{\theta} \log p_{\mathbf{\theta}}^{\rm R}(\mathbf{k}^t | \mathbf{c}^t) + \nabla_{\upsilon} V_{\upsilon}(\mathbf{v})]\cdot \delta \big],
\end{split}
\end{eqnarray}
where $\delta$ is the td-error defined as $\delta= r^{t+1} + \gamma V_{\upsilon^{-}}(\mathbf{v}) - V_{\upsilon}(\mathbf{v})$.
The algorithm is implemented with deep learning platform
{\fontsize{8}{5}\sffamily PaddlePaddle}\footnote{\url{https://github.com/PaddlePaddle/Paddle}}.
We train the network with Adagrad~\citep{Adagrad} with a batch size of $16$ and a learning rate of $\displaystyle\num{1e-5}$. Discount factor $\gamma=0.99$.
Experience replay is used in practice~\citep{DQN}.

\section{Experimental Results}
\label{sec:Exp}

We evaluate the performance of the proposed approach under several different settings to demonstrate its ability of interactive language learning. 
For training efficiency, we construct a simulated environment for language learning as shown in Figure~\ref{fig:example}.
We consider the case with four different objects around the learner in each direction (\emph{S, N, E, W}), which are randomly sampled from a set of objects for each session. 
Within this environment, a teacher interacts with the agent about objects around in three different forms: (1)~asking a question as ``\emph{what is on the south}'', ``\emph{where is apple}'' and the agent answers the question; (2)~describing objects around as ``\emph{apple is on the east}'' and agents repeat the statement; (3)~saying nothing (``.'') then agent describes objects around and gets a feedback from teacher. The agent receives a positive reward (\scalebox{0.9}[1]{$\textstyle r\!\!=\!\!+1$}) if it behaves correctly (generates a correct answer to a question from teacher or produces a correct statement if teacher says nothing) and a negative reward (\scalebox{0.9}[1]{$\textstyle r\!\!=\!\!-1$}) otherwise. Reward is used to represent teacher's non-verbal feedback such as \emph{nodding} as a form of encouragement.
Besides reward feedback, teacher also provides a verbal feedback including the expected answer in the form of ``\emph{X is on the east}'' or ``\emph{on the east is X}'' and with prefix (``\emph{yes/no}'') added with a probability of half. 
The speaking action from the agent is correct if it outputs a sentence matches exactly with the expected answer in one of the above forms. There is a possibility for the learner to generate a new correct sentence that beyond teacher's knowledge. This is not handled in our current work due to the usage of a scripted teacher.

\begin{figure}[t]
	
	\begin{minipage}[t]{0.42\linewidth}
		\vspace{0pt}
		\begin{overpic}[viewport = 12 2 550 400, clip, width =6cm]{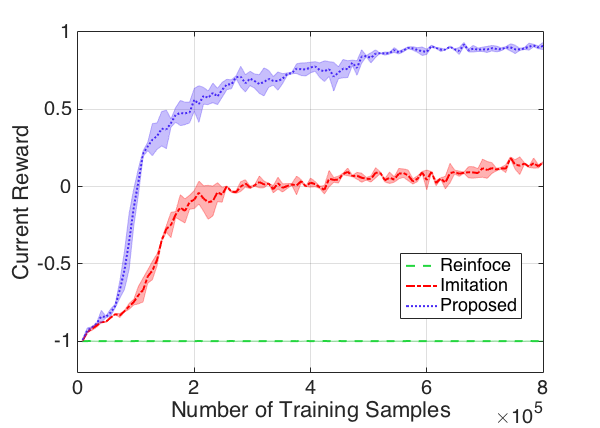}
			\put(2,2){\sffamily \scriptsize {\textcolor{black}{{(a)}}}}
			\put(145,2){\sffamily \scriptsize {\textcolor{black}{{(b) Quantative Resutls}}}}		
			
		\end{overpic}
	\end{minipage}
	\vspace{-0.05in}
	\hspace{0.2in}
	\begin{minipage}[t]{0.48\linewidth}
		\captionof{table}{\footnotesize Testing Results with Mixed Config.} 
		\label{tab:mixed_setting}
		\vspace{-0.1in} 
		\centering
		\footnotesize
		\renewcommand{\arraystretch}{1.1}%
		\begin{tabular}[t]{ |c|c|c|c| }
			\hline	
			Settings&  Reinforce & Imitation & Proposed\\
			\hline\hline
			{\scriptsize Compositional-gen.}&  0.0\%  & 83.7\%    & 98.9\% \\[0.ex] \hline
			{\scriptsize Knowledge-transfer}&  0.0\%  & 81.6\%    & 97.5\%\\ \hline 			
		\end{tabular}
		
		\vspace{0.15in}
		\captionof{table}{\footnotesize Testing Results wtih Held-out Config.} \label{tab:held_out_setting} 
		\vspace{-0.1in}			
		\centering
		\footnotesize	
		\begin{tabular}[t]{ |c|c|c|c| }
			\hline
			Settings&   Reinforce & Imitation & Proposed\\
			\hline\hline
			{\scriptsize Compositional-gen.}   &  0.0\%  & 75.1\%    & 98.3\%\\ \hline
			{\scriptsize Knowledge-transfer} &  0.0\%  & 70.4\%   & 89.0\%\\ \hline 
		\end{tabular}
	\end{minipage}
	\vspace{0.1in}
	\caption{\textbf{Evaluation results.} (a)~Evolution of reward during training. (b)~Comparison of the proposed approach with  {Reinforce} and {Imitation} approaches across different settings and configurations. \emph{Mixed config} denotes the configuration involving interactions with all objects.  \emph{Held-out config} denotes the configuration involving interactions with only the objects that are inactive during training.}
	\label{fig:reward_curve}
\end{figure}

\begin{figure}[t]
	\hspace*{-0.0cm} 
	\begin{overpic}[viewport = 39 39 242 241, clip, width = 2cm]{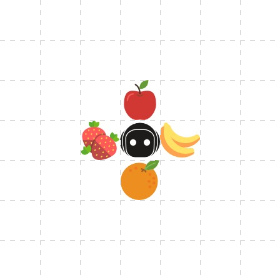}
		\put(-5,10){\sffamily \tiny {\textcolor{black}{{T: \emph{what is on the north}}}}}
		\put(-5,0){\sffamily \tiny {\textcolor{black}{{L: \emph{on the north is apple}}}}}
		\put(-5,-10){\sffamily \tiny {\textcolor{black}{{T: \emph{yes  apple is on the north} {\color{blue}[$+$]}}}}}
		\put(85,70){\includegraphics[viewport = 39 39 241 241, clip, width = 0.8cm]{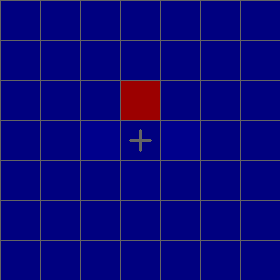}}
		\put(85,59){{\sffamily \scalebox{.6} {\textcolor{black}{{att. map}}}}}
		\put(-3,90){\sffamily \scriptsize {\textcolor{black}{{(a)}}}}
	\end{overpic}
	\hspace{1.5cm}
	\begin{overpic}[viewport = 39 39 242 241, clip, width = 2cm]{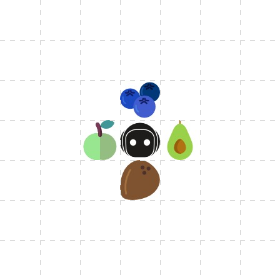}
		\put(-5,10){\sffamily \tiny {\textcolor{black}{{T: \emph{what is on the east}}}}}
		\put(-5,0){\sffamily \tiny {\textcolor{black}{{L: \emph{on the east is avocado}}}}}
		\put(-5,-10){\sffamily \tiny {\textcolor{black}{{T: \emph{avocado is on the east} {\color{blue}[$+$]}}}}}
		\put(85,70){\includegraphics[viewport = 39 39 241 241, clip, width = 0.8cm]{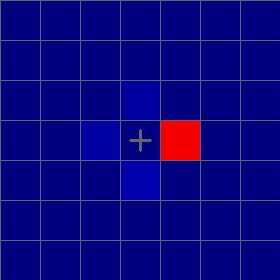}}
		\put(85,59){{\sffamily \scalebox{.6} {\textcolor{black}{{att. map}}}}}
		\put(-3,90){\sffamily \scriptsize {\textcolor{black}{{(b)}}}}
	\end{overpic}
	\hspace{1.5cm}
	\begin{overpic}[viewport = 39 39 242 241, clip, width = 2cm]{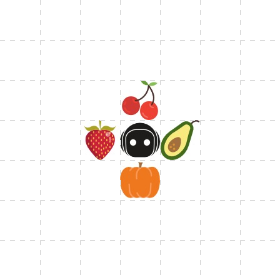}
		\put(-10,10){\sffamily \tiny {\textcolor{black}{{T: \emph{where is strawberry}}}}}
		\put(-10,0){\sffamily \tiny {\textcolor{black}{{L: \emph{strawberry is on the west}}}}}
		\put(-10,-10){\sffamily \tiny {\textcolor{black}{{T: \emph{yes strawberry is on the west} {\color{blue}[$+$]}}}}}
		\put(85,70){\includegraphics[viewport = 39 39 241 241, clip, width = 0.8cm]{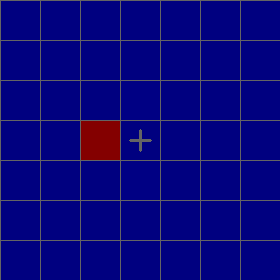}}
		\put(85,59){{\sffamily \scalebox{.6} {\textcolor{black}{{att. map}}}}}
		\put(-3,90){\sffamily \scriptsize {\textcolor{black}{{(c)}}}}
	\end{overpic}
	\hspace{1.5cm}
	\begin{overpic}[viewport = 39 39 242 241, clip, width = 2cm]{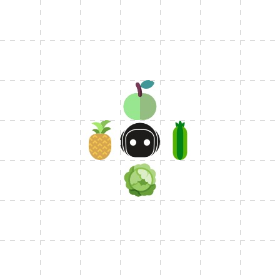}
		\put(-10,10){\sffamily \tiny {\textcolor{black}{{T: .}}}}
		\put(-10,0){\sffamily \tiny {\textcolor{black}{{L: \emph{on the east is cucumber}}}}}
		\put(-10,-10){\sffamily \tiny {\textcolor{black}{{T: \emph{yes on the east is cucumber} {\color{blue}[$+$]}}}}}
		\put(85,70){\includegraphics[viewport = 39 39 241 241, clip, width = 0.8cm]{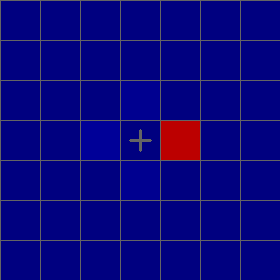}}
		\put(85,59){{\sffamily \scalebox{.6} {\textcolor{black}{{att. map}}}}}
		\put(-3,90){\sffamily \scriptsize {\textcolor{black}{{(d)}}}}
	\end{overpic}
	\vspace{0.15in}
	\caption{\textbf{Example results.} (a-b) \emph{what} questions. (c)~\emph{where} question. (d)~teacher says nothing (``.'') and the agent is expected to produce a statement.
		For each example, we show the visual image, the conversion dialogues between teacher and learner, as well as the \emph{attention map} ({\sffamily att. map}) generated from the learner when producing the response to teacher (overlaid on top-right).
		The attention map is rendered as a heat map, with red color indicating large value while blue indicating small value.
		Grid lines are overlaid on top of the attention map to for visualization purpose. The position of the learner is marked with a cross in the attention map
		({{\small\sffamily T}/{\small\sffamily L}}: teacher/learner, {\small[$\color{blue}{+}$/$\color{red}{-}$]}: positive/negative rewards).
	}
	\label{fig:example_with_att}
	\vspace{-0.15in}
\end{figure}

\textbf{Language Learning Evaluation.} We first validate the basic language learning ability of the proposed approach under the interactive language learning setting. In this setting, the teacher first generates a sentence for the learner, then the learner will respond, and the teacher will provide feedback in terms of sentence and reward.
We compare the proposed approach with two baseline approaches: (1)~\textbf{Reinforce} which uses directly reinforcement for learning from teacher's reward feedback~\citep{Sutton}; 
(2)~\textbf{Imitation} which learns by mimicking teacher's behavior~\citep{Seq2Seq}. 
Experimental results are shown in Figure~\ref{fig:reward_curve}. 
It is interesting to note that learning directly from reward feedback only (\textbf{Reinforce}) does not lead to successful language acquisition. This is mainly because of the low possibility of generating a sensible sentence by random exploration, and the even lower possibility of generating the correct sentence, thus the received reward can stay at $-1$.
On the other hand, the \textbf{Imitation} approach  performs better than \textbf{Reinforce}, due to the \emph{speaking} ability it gained through mimicking.
The proposed approach achieves reward higher than both compared approaches, 
due to the effectiveness of the joint formulation, which can fully leverage the feedback signals appeared naturally during conversion for learning.
This indicates the effectiveness of the proposed approach for language learning under the interactive setting.
Similar behaviors have been observed during testing.
We further visualize some examples as shown in Figure~\ref{fig:example_with_att} along with the generated attention maps. As can be observed from the results, the proposed approach can successfully generate correct attention maps for both \emph{what} and \emph{where} questions. When teacher says nothing (``.''), the agent can generate a statement describing an object around correctly.

\vspace{1.in}

\textbf{Zero-shot Dialogue.}
An intelligent agent is expected to have an ability to generalize.
While this is {not the main focus} on the paper, we use it as a way to assess the language learning ability of an approach.
Experiments are done in following two settings.
(1)~\textbf{Compositional generalization}: the learner interacts with the teacher about objects around during training, but does not have any interaction with certain objects (referred to as \emph{inactive} objects) at particular locations, while in testing the teacher can  ask questions about an object regardless of its location. 
It is expected that 
a good learner should be able to generalize the concepts it learned  about both \emph{objects} and \emph{locations}  as well as the acquired conversation skills and can interact successfully in natural language with teacher about novel \{\emph{object}, \emph{location}\} combinations that it never experienced before. 
(2)~\textbf{Knowledge transferring}: teacher asks learner questions about the
objects around. For certain objects, the teacher only provides descriptions without asking questions during training, while in testing, the teacher can ask questions about any object present in the scene.
The learner is expected to be able to 
transfer the knowledge learned from teacher's description to generate an answer to teacher’s question about these objects.
Experiments are carried out under these two settings for two configurations (\emph{mixed} and \emph{held-out}) and experimental results are summarized in Table~\ref{tab:mixed_setting} and Table~\ref{tab:held_out_setting} respectively.
\emph{Mixed configuration} denotes the case in which a mixture of interactions with all objects regardless of whether they are active or inactive during training.  \emph{Held-out configuration} denotes the case involving interactions with only the objects that are inactive during training.
The results shows that the \textbf{Reinforce} approach performs poorly under both settings due to the lack of basic language-related abilities as mentioned in the previous section.
The \textbf{Imitation} approach performs better than \textbf{Reinforce} mainly due to its language speaking ability through mimicking.
Note that the held-out configuration is a subset of the mixed-configuration involving only novel objects/combinations, thus is more difficult than the mixed case.
It is interesting to note that the proposed approach maintains a consistent behavior under the more difficult held-out configuration and  outperforms the other two approaches under both settings, demonstrating its effectiveness in interactive language learning.

\section{Conclusion}
\label{sec:con}
We present an interactive  setting for grounded natural language learning and propose an approach that achieves effective interactive natural language learning by fully leveraging the feedbacks that arise naturally during interaction through joint imitation and reinforcement.
Experimental results show that the proposed approach provides an effective way for natural language learning in the interactive setting and enjoys desirable generalization and transferring abilities under several different scenarios.
As for future work, we would like to explore the direction of explicit modeling of learned knowledge~\citep{DYang2003} and fast learning about new concepts~\citep{L2L}.
Another interesting direction is to connect the language learning task presented in this paper with other heterogeneous tasks such as navigation.

\subsubsection*{Acknowledgements}
We thank Xiaochen Lian, Zhuoyuan Chen, Yi Yang and Qing Sun for their discussions and comments.

{
	\setlength{\bibsep}{0.1pt plus 0.3ex}
	\bibliographystyle{abbrvnat}
	\footnotesize
	\bibliography{L2T}
}
\end{document}